\documentclass[10pt,twocolumn,letterpaper]{article}

\usepackage{cvpr}
\usepackage{times}
\usepackage{epsfig}
\usepackage{graphicx}
\usepackage{amsmath}
\usepackage{amssymb}

\usepackage[american]{babel}
\usepackage{algorithmic}
\usepackage{algorithm}
\usepackage{threeparttable}
\usepackage{multirow}

\newcommand\norm[1]{\left\lVert#1\right\rVert}
\renewcommand{\vec}[1]{\boldsymbol{#1}} % Uncomment for BOLD vectors.

\usepackage{enumitem}
\setenumerate[1]{itemsep=0pt,partopsep=0pt,parsep=\parskip,topsep=0pt}
\setitemize[1]{itemsep=0pt,partopsep=0pt,parsep=\parskip,topsep=0pt}
\setdescription{itemsep=0pt,partopsep=0pt,parsep=\parskip,topsep=0pt}
\usepackage[pagebackref=true,breaklinks=true,letterpaper=true,colorlinks,bookmarks=false]{hyperref}

\cvprfinalcopy % *** Uncomment this line for the final submission

\def\cvprPaperID{2357} % *** Enter the CVPR Paper ID here

\begin{document}

%%%%%%%%% TITLE
\title{AutoPruner: An End-to-End Trainable Filter Pruning Method for \\Efficient Deep Model Inference}

%% omit the following lines up until the closing ``}''.
%% Additional authors and addresses can be added with ``\and'',
%% just like the second author.
%% To save space, use either the email address or home page, not both
%\and

\author{Jian-Hao Luo \qquad Jianxin Wu\\
	National Key Laboratory for Novel Software Technology \\
	Nanjing University, Nanjing, China\\
	{\tt\small luojh@lamda.nju.edu.cn, wujx2001@nju.edu.cn} 
}

\maketitle
%\thispagestyle{empty}

%%%%%%%%% ABSTRACT
\begin{abstract}
	Channel pruning is an important family of methods to speed up deep model's inference. Previous filter pruning algorithms regard channel pruning and model fine-tuning as two independent steps. This paper argues that combining them into a single end-to-end trainable system will lead to better results. We propose an efficient channel selection layer, namely AutoPruner, to find less important filters automatically in a joint training manner. Our AutoPruner takes previous activation responses as an input and generates a true binary index code for pruning. Hence, all the filters corresponding to zero index values can be removed safely after training. We empirically demonstrate that the gradient information of this channel selection layer is also helpful for the whole model training. By gradually erasing several weak filters, we can prevent an excessive drop in model accuracy. Compared with previous state-of-the-art pruning algorithms (including training from scratch), AutoPruner achieves significantly better performance. Furthermore, ablation experiments show that the proposed novel mini-batch pooling and binarization operations are vital for the success of filter pruning.
\end{abstract}

\section{Introduction}
Deep neural networks suffer from serious computational and storage overhead. For example, the VGG16~\cite{VGG16} model, which has 138.34 million parameters, requires more than 30.94 billion floating-point operations (FLOPs) to recognize a single $224\times 224$ input image. It is impossible to deploy such cumbersome models on real-time tasks or resource constrained devices like mobile phones~\cite{ThiNet}.

To address this problem, many model compression or acceleration algorithms have been proposed~\cite{Han15NIPS, ThiNet, sparse_selection, Chen15ICML, Lebedev16CVPR, Denton14NIPS, Denil13NIPS, Wu16CVPR}. Among these methods, pruning is an important direction. By removing unimportant neurons, the model size and computational cost can be reduced. We can roughly divide pruning approaches into 3 categories: \emph{connection level}, \emph{filter level} and \emph{layer level} pruning. A simple method is to discard connections according to the magnitude of their weight values~\cite{Han15NIPS}. However, such an unconstrained pruning strategy will lead to an irregular network structure. It may slow down the actual inference speed even though the sparsity is high~\cite{Wen16NIPS}. Hence, structured pruning, such as filter level pruning~\cite{RNP, ThiNet, Sunjian_ICCV17, JianguoLi_ICCV17}, is attracting more and more attentions in recent years. In filter level pruning, the whole filter will be removed if it is less important. Hence, the original network structure will not be damaged after pruning.

\begin{figure}
	\centering
	\includegraphics[width=1.0\linewidth]{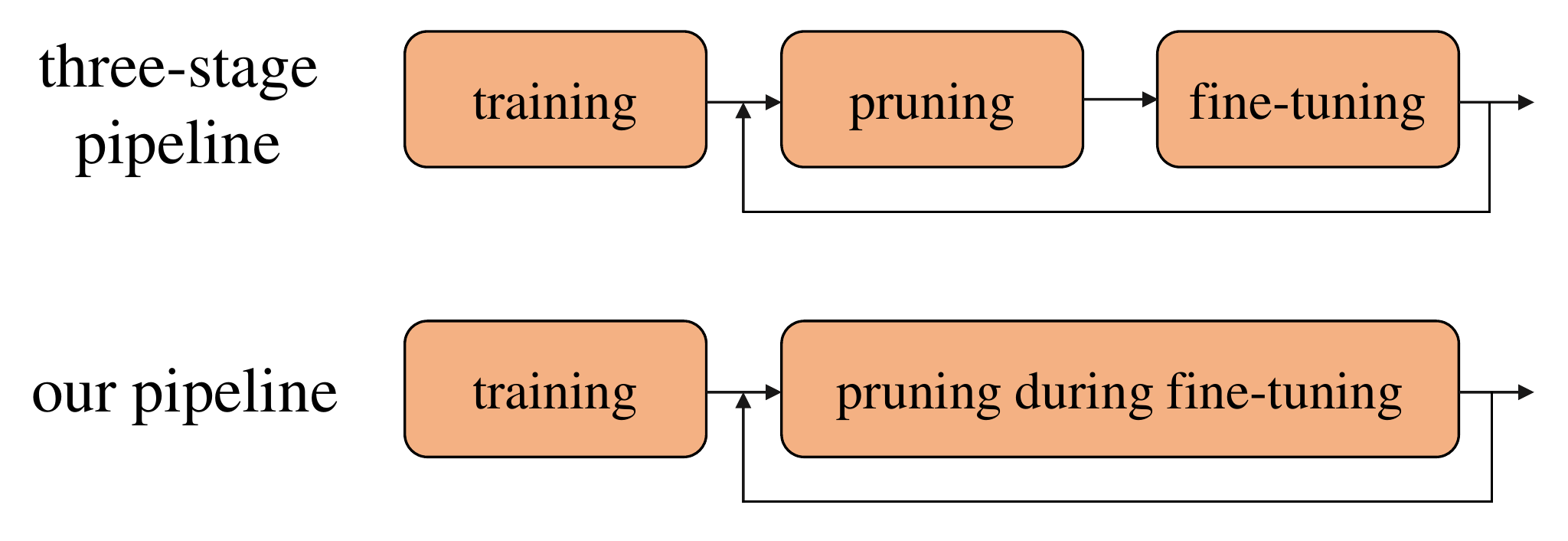}
	\caption{Overview of network pruning pipeline. The first row is a typical three-stage pruning pipeline, which regards pruning and fine-tuning as two independent processing steps. In the proposed AutoPruner method, we integrate filter selection into model fine-tuning. During fine-tuning, our method will gradually erase unimportant filters in an automatic manner.}
	\label{three_stage}
\end{figure}

As illustrated in the first row of Figure~\ref{three_stage}, most current filter pruning methods adopt a three-stage pipeline. Starting from a pre-trained model, they try to find a better evaluation criteria for measuring the importance of filters, discard several weak filters, then fine-tune the pruned model to recover its accuracy. However, it is hard to find a perfect criterion that can work well on all networks and tasks. More importantly, pruning and model training are two \emph{independent} processing steps in this pipeline. Hence, here comes an interesting question: \emph{could fine-tuning be utilized to guide the selection of weak filters}? In other words, \emph{can we teach the model to decide which filter should be pruned}?

In order to answer this question, we propose a novel end-to-end trainable method, namely AutoPruner, to explore a new way for CNN pruning. By integrating filter selection into model training, the fine-tuned network can select unimportant filters automatically. Our AutoPruner can be regarded as a new CNN layer, which takes the activation of previous layer as an input, and generate a unique \emph{binary} code. A 0 value in the binary code means its corresponding filter's activation will always be 0, hence can be safely eliminated. And ``unique'' means our AutoPruner is a static method, all the zero filters will be removed forever.

Experimental results on fine-grained CUB200-2011 dataset~\cite{CUB200} and the large-scale image recognition task ILSVRC-2012 dataset~\cite{ImageNet} have demonstrated the effectiveness of the proposed AutoPruner. AutoPruner outperforms previous state-of-the-art approaches with a similar or even higher compression ratio. We also compared AutoPruner with a simple but powerful method: training from scratch. The result of this experiment reveals that our AutoPruner achieves better accuracy, which is really useful to obtain a more accurate small model. 

The key advantages of AutoPruner and our contributions are summarized as follows.
\begin{itemize}
	\item \textbf{End-to-end trainable in a single model.} Filter selection and model fine-tuning are integrated into a single end-to-end trainable framework. We empirically demonstrate that these two processing steps can promote each other. The model will select better filters automatically during fine-tuning. And, the gradients of filter selection are also helpful to guide the training of previous convolution layers. In other words, \emph{fine-tuning can be utilized to guide pruning, and gradually erasing weak filters (\ie, pruning) is really important to obtain a more accurate model (\ie, fine-tuning)}.
	\item \textbf{Adaptive compression ratio and multi-layer compression.} We propose a novel loss function to ensure the sparsity of binary code could converge to a predefined compression ratio. But we encourage network to determine the actual sparsity by itself, which will take both accuracy and compression ratio into consideration. And, we can compress multiple layers simultaneously to reduce training cost.
	\item \textbf{Good generalization ability.} The proposed method achieves better performance on multiple datasets and networks compared with previous state-of-the-art algorithms. Our method is easy to implement and can be extended to other deep learning libraries.
\end{itemize}

\begin{figure*}
	\centering
	\includegraphics[width=0.95\linewidth]{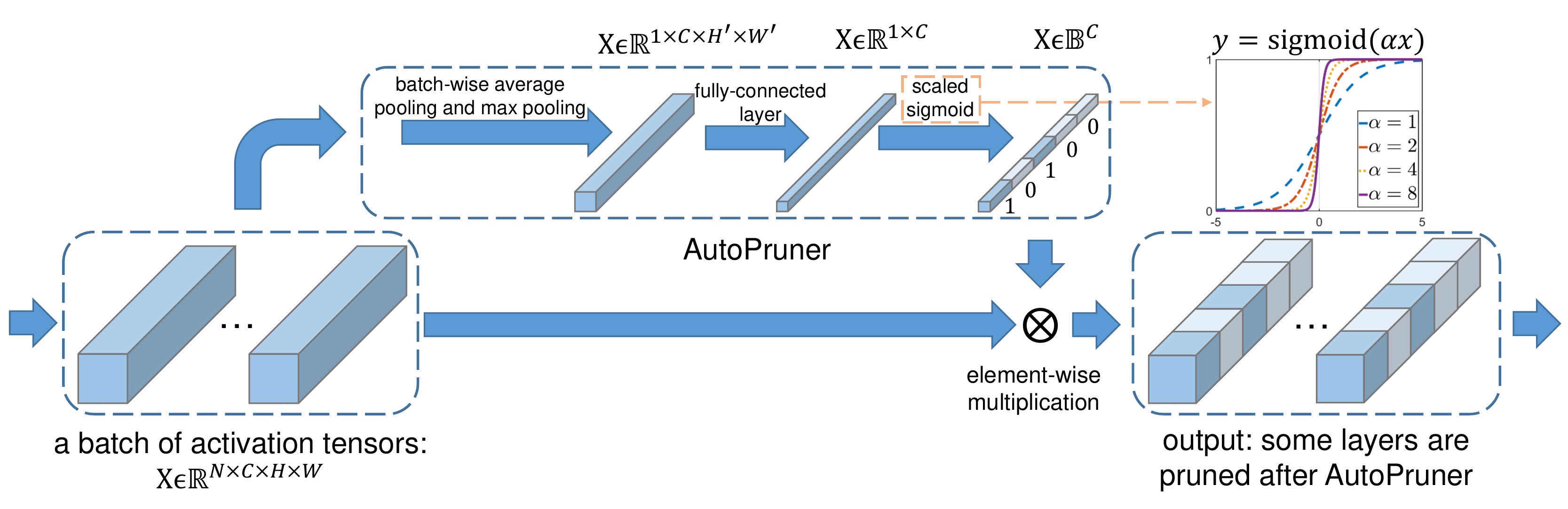}
	\caption{Framework of the proposed AutoPruner layer. Given a mini-batch of activation tensors, we use a new batch-wise average pooling and a standard max pooling to generate a single tensor. This tensor is projected into a C-dimensional vector via a fully-connected layer, where C is the number of channels. Finally, a novel scaled sigmoid function is used to obtain an approximate binary output. By gradually increase the value of $\alpha$ in scaled sigmoid function, the output of AutoPruner will gradually become a C-dimensional binary code. After training, all the filters and channels corresponding to the zeros index values will be pruned away to obtain a smaller and faster network. The new added AutoPruner layer will be removed too.}
	\label{framework}
\end{figure*}
\section{Related Work}
Pruning is a classic method to reduce model complexity~\cite{Han15NIPS, Wen16NIPS, Louizos17NIPS, RNP, Alvarez17NIPS, Aghasi17NIPS, ThiNet}. Compared with training the same structure from scratch, pruning a pretrained redundant model achieves much better results~\cite{ThiNet, Li17ICLR}. This is mainly because of the highly non-convex optimization nature in model training. And, certain level of model redundancy is necessary to guarantee enough capacity during training. However, such a cumbersome model will slow down the running speed of model inference. And the model capacity is also too large when transfer to a much smaller dataset. Hence, there is a great need to remove the redundancy.

The most intuitive idea to evaluate neuron importance is based on the magnitude of its weight value. Han \etal~\cite{Han15NIPS, Han16ICLR} proposed an iterative pruning method to discard small-weight connections which are below a predefined threshold. However, connection level pruning can lead to an irregular convolution, which needs a special algorithm or dedicated hardware for efficient inference, thus is hard to harvest actual computational savings. To address the weakness of non-structured random pruning, some structured sparsity learning algorithms have been proposed~\cite{Lebedev16CVPR, Wen16NIPS, Wen18ICLR}. In these works, only groups of structured neurons, such as the whole channel or filter, will be pruned.

Recently, filter level pruning has drawn a significant amount of interests from both academia and industry. Luo~\etal~\cite{ThiNet} formally established filter pruning as an optimization problem, and removed the less important filters based on the statistics of next layer. Similarly, He~\etal~\cite{Sunjian_ICCV17} proposed a LASSO regression based method to select unimportant channels. Liu~\etal~\cite{JianguoLi_ICCV17} introduced channel scaling factors to denote the importance of each layer. Yu~\etal~\cite{NISP} propagated the importance scores of final responses to every neuron and formulate network pruning as a binary integer optimization problem. All of these methods are trying to find a better importance evaluation method.

However, these three-stage pruning algorithms all regarded channel (or filter) selection and model fine-tuning as two separate steps. We argue that combining them into a single end-to-end system will be a better choice: \emph{the information flowed from uncompressed layers can be used to guide the pruning of current layer}. 

There are some explorations to prune networks beyond the three stage framework, too. Lin~\etal~\cite{RNP} introduced a novel dynamic pruning method based on reinforcement learning. The network is dynamically pruned according to the output Q-value of a decision network. By contrast, our method is static, the zero filters will be removed forever. He~\etal~\cite{AMC} introduced AutoML for model pruning. They leverage reinforcement learning to efficiently sample the network design space. Huang~\etal~\cite{sparse_selection} adopted scaling factors to indicate the importance of each neuron, and formulate it as a joint sparse regularized optimization problem. Their method is very similar to ours. The major difference is whether pruning information will participate in the training of previous layers or not. We empirically demonstrate that \emph{the gradient of channel selection layer is also helpful for model training}.

\section{Automatic Filter Pruner}

In this section, we propose our adaptive end-to-end trainable filter pruning method: AutoPruner. We will give a comprehensive introduction to the AutoPruner pipeline as well as several important implementation details.

\subsection{The Proposed AutoPruner Pipeline}

Figure~\ref{framework} shows the framework of AutoPruner. AutoPruner can be regarded as an independent layer, whose input is the responses (after activation function) of a standard convolution layer. An approximate binary index code is generated by AutoPruner. We then use element-wise multiplication to combine it with activation tensors. By gradually forcing the scaled sigmoid function to emit binary index codes, some channels in the activation tensors will gradually become all zero (\ie, they will be erased gradually). Hence, we can safely pruned these channels (or filters) away. Since AutoPruner is trained in an end-to-end manner, channel selection and model fine-tuning are combined together, and will promote each other during training.

After training, the binary index code is used for filter pruning. All the filters in previous layer and all the channels in the filters of the next layer will be removed if their corresponding index value is 0. The new added AutoPruner layer will be removed too. Hence, the pruned model has no difference in model structure with previous pruning method.

Next, we will go into details about the proposed AutoPruner. Our method consists of three major parts: pooling, coding and binarization. We will give a comprehensive introduction to them separately.

\subsubsection{Pooling} \label{sec_pooling}

We use $\text{X} \in \mathbb{R}^{N \times C\times H \times W}$ to denote the activation output of a convolution layer, with mini-batch size $N$, $C$ channels, $H$ rows and $W$ columns. First, batch-wise average pooling is used to aggregate all the elements among different images:
\begin{equation}
\text{X}' = \frac{1}{N}\sum_{i=1}^{N}X_{i, :, :, :}.
\end{equation}

Note that \emph{the generated index code should be decided by the layer, not one example}. In other words, we need the activations, which is computed from different images, to be transformed into a unique index code in one specific layer. The batch-wise average pooling mixes information of different images and is helpful in achieving the consistency of index codes among different images. The subsequent binarization technique in AutoPruner enable us to generate a unique index code for each layer.

Next, the pooled tensor $\text{X}'$ is fed into a standard max-pooling function with $2\times 2$ filter size and stride 2 to reduce memory consumption. We find this reduction of information does not affect model accuracy obviously, but adding this step will save GPU memory consumption and training time in the coding stage. However, as shown later, a large spatial pooling operation (such as global average pooling) is harmful.

\subsubsection{Coding} \label{sec_coding}

Then, in the coding stage, the pooled tensor will be projected into a C-dimensional vector. A fully-connected layer, whose weights are denoted as $\widehat{\mathcal{W}} \in \mathbb{R}^{C \times (CH'W')}$, is used here to generate a C-dimensional vector, where C is the channel number in input activations. 

Initialization plays an important role in model training. We tried the MSRA method~\cite{MSRA_Init} to initialize our fully-connected weights. However, we find this approach is not suitable. The index code must converge to 0-1 binary values after training (cf. the next section), which means the variance of the initial weights could not be too small. Hence, we propose a new strategy, which increases the value of standard deviation by $10\times$. In our method, each weight in the coding fully-connected layer is initialized with a zero-mean Gaussian distribution whose standard deviation is
\begin{equation}
10\times \sqrt{\frac{2}{n}}.
\end{equation}
Here, $n=C\times H' \times W'$ is the number of input elements.

\subsubsection{Binarization} \label{subsubsec: binary}

To the best of our knowledge, we are the first to require that the generated index code should be 0-1 values, which is essential in assuring the pruning quality. Combined with the mini-batch pooling operation, binarization ensures that all the examples in a mini-batch could be transformed to the same unique index code eventually. 

We use a scaled sigmoid function to generate the approximate binary code:
\begin{equation} \label{scaled_sigmoid}
y=\text{sigmoid}(\alpha x),
\end{equation}
where $\alpha$ is a hyper-parameter, which controls the magnitude of output values. As illustrated in the top right corner of Figure~\ref{framework}, by gradually increasing the value of $\alpha$, scaled sigmoid function can generate approximate binary code. When $\alpha$ is large enough, the approximate binary values will become 0-1 values eventually. In others words, pruning is finished during fine-tuning, and which filter should be pruned away is totally decided by network itself.

Such a gradual binarization strategy is helpful for obtaining a more accurate model. When some channels are becoming smaller (0.5 $\to$ 0), the corresponding filters will stop updating gradually. At the same time, other channels are becoming larger (0.5 $\to$ 1), which will force the network to pay more attention to the preserved filters.

Another major benefit of binarization is that pruning and fine-tuning is now seamlessly integrated together. Pruning can be finished during mode fine-tuning. If the binary code for one channel is 0, we know its activation values will always be 0 for any input image. And, If the binary code for one channel is 1, the activation values do not change in the element-wise multiplication operation. Hence, after the codes become binary, \emph{removing all pruning block and the pruned filters will \emph{not} change the network's prediction}.

\subsection{Sparsity Control and Loss Function} \label{subsec:sparsity}

So far, we have introduced the whole pipeline of AutoPruner. The next question is, can we control the sparsity of the output index code? In some real-world application scenarios, the inference speed or model size is constrained. For example, a scene segmentation network should return predictions within 50 ms in self-driving vehicles for safety consideration. These constraints can be solved via a predefined compression rate.

To address this problem, we proposed a simple yet efficient sparsity control regularized loss function. We use $\textbf{v}$ to denote the index code vector generated by AutoPruner. One of the most commonly used sparsity regularization method is the convex relaxation $\ell_1$-norm, which is defined by $\norm{\textbf{v}}_1$. However, $\ell_1$-norm can not control the sparsity to an expected value. Noticing that $\textbf{v}$ is an approximate binary vector, we can use $\frac{\norm{\textbf{v}}_1}{C}$ to denote the percentage of 1 approximately, \ie, the percentage of preserved filters. Here, $C$ is the length of vector $\textbf{v}$.

Given a predefined compression rate $r\in [0, 1]$ (the percentage of preserved filters), we can formulate our loss function as:
\begin{equation}
\min \mathcal{L}_{\text{classification}} + \lambda \norm{\frac{\norm{\textbf{v}}_1}{C} - r}_2^2.
\end{equation} 
The first term is a standard classification loss (\eg, cross-entropy loss), the second one controls the model compression rate, and $\lambda$ balances the relative importance between these two terms. Furthermore, its value is adaptively adjusted according to current compression ratio:
\begin{equation}
\lambda = 100 \times |r_b - r|,
\end{equation}
where $r_b$ is the current compression ratio. In practice, we collect several values of $\textbf{v}$ during model training, and calculate the current compression ratio $r_b$ based on these data. ``100'' converts percentage values to a normal one. $\lambda$ is initialized by 10, and adaptively changed during model training. If current compression ratio is far from our expected goal, $\lambda$ is relatively large. Hence, the model could pay more attention to change the sparsity of code index $\textbf{v}$. Once we have got the expected $\textbf{v}$, $\lambda$ will finally become 0, which means the network can focus on classification task.

We want to emphasize that the actual compression rate is determined by the network itself. Our novel loss function can control the sparsity, but the actual value can still vary. For example, if we want to prune half of the filters and set $r=0.5$. After training, the actual value of $r_b$ may be 0.52 or 0.48. Hence, the proposed AutoPruner can achieve an \emph{adaptive} network compression.

And, we can repeat the operation in Figure~\ref{framework} for multiple layers and compress them simultaneously.

\subsection{Initialization and Binarization Control} \label{sec_implementation}

As we have emphasized in Sec.~\ref{sec_coding}, initialization is essential in our framework. The initial value of code index $\textbf{v}$ is determined by three factors: input tensor, weight value and $\alpha$. In Sec.~\ref{sec_coding}, we have introduced the initialization method for fully-connected weights. As for the input tensor, its magnitude is uncertain in different layers. Hence, we can only adjust the value of $\alpha$ to control initial value of $\textbf{v}$.

There are a few observations that are worth discussing.
\begin{itemize}
	\item When $\alpha$ is too large, $\textbf{v}$ will become binary quickly. In this case, the discard filters are determined before training. Hence, AutoPruner will degenerate into random selection.
	\item When $\alpha$ is too small, $\textbf{v}$ may be difficult to or even impossible to converge into binary values. What is more troubling is that $\textbf{v}$ can be stuck in small values, \ie, all the elements are smaller than 0.5, but can never be pulled back around 1.
	\item Unfortunately, appropriate value of $\alpha$ can differ greatly in different layers due to the magnitude of input. It is impossible to find an appropriate $\alpha$ that can work well for all layers.
\end{itemize}
Based on the above observations, we propose an efficient adjustment scheme to find an approximate value for $\alpha$. As we have discussed in section~\ref{subsubsec: binary}, we should increase $\alpha$ gradually. Starting from the initial value $\alpha_{start}$, we linearly increase its value and finally stop at $\alpha_{stop}$. Hence, this question is equivalent to find the values for $\alpha_{start}$ and $\alpha_{stop}$.

The first step is to find $\alpha_{stop}$ which can produce a binary output in scaled sigmoid function (Eq.~\ref{scaled_sigmoid}). However, this value varies greatly in different networks. Our method is to test several numbers until its outputs are all 0-1 values. Note that, this step can be finished quickly before model fine-tuning: we only need to try several numbers. For example, in our internal test we find $\alpha_{stop}=2$ is enough for VGG16~\cite{VGG16}. But for ResNet-50~\cite{ResNet}, $\alpha_{stop}$ should not be smaller than 100. 

As for $\alpha_{start}$, it is a hyper-parameter. We heuristically set it to 0.1 for VGG16, and 1 for ResNet-50. In order to avoid the influence of small $\alpha$, we adopt a simple yet efficient method. We will check the values of $\textbf{v}$ after several epochs (\eg, 2 or 3 epochs). If it is still far from convergence, $\alpha$ will be increased faster, forcing it converge into binary values. Using such a simple strategy, our AutoPruner model can generate a unique binary code successfully.

\section{Experimental Results}

In this section, we will empirically study the benefits of our AutoPruner method. We compared our method with other state-of-the-art pruning approaches on two standard datasets: CUB200-2011~\cite{CUB200} and ImageNet ILSVRC-12~\cite{ImageNet}. Two widely used deep models, VGG16~\cite{VGG16} and ResNet-50~\cite{ResNet}, were pruned. All the experiments were conducted using pyTorch on M40 GPUs.

\subsection{FLOPs Computation}
FLOPs, namely floating-point operations, is a popular metric to evaluate the complexity of CNN models. Following the setting of~\cite{Molchanov17ICLR}, the FLOPs in convolutional layers is calculated by:
\begin{equation} \label{eq:FLOPs}
\text{FLOPs}=2HW(C_{in}K^2+1)C_{out},
\end{equation}
where $H$, $W$, $C_{out}$ is the height, width and channel number of output tensor, $K$ is the kernel size, $C_{in}$ refers to the number of input channels, and $1$ means the FLOPs in bias term. Note that, we regard a single vector multiplication as two floating-point operations (multiplication
and addition). However, in some papers~\cite{ResNet, sparse_selection}, it may be regarded as one FLOP. For a fair comparison, we will re-calculate the FLOPs number if it is not computed by Eq.~\ref{eq:FLOPs}. 

\subsection{CUB200-2011} \label{exp_cub}

\begin{table*}
	\caption{Compressing VGG16 on CUB200-2011 dataset using different algorithms and compression rates. For AutoPruner, we run it 3 times and report the mean$\pm$std values (\%).}
	\label{vgg16_CUB}
	%\footnotesize
	%\setlength{\tabcolsep}{10pt}
	\centering
	\begin{threeparttable}
		\begin{tabular}{|p{5cm}|c|c|c|c|c|c|}
			\hline
			\multirow{2}{*}{Method} & \multicolumn{3}{c|}{compression rate $r=0.5$} & \multicolumn{3}{c|}{compression rate $r=0.2$} \\
			\cline{2-7} & top-1 (\%) & top-5 (\%) & \#FLOPs & top-1 (\%) & top-5 (\%) & \#FLOPs \\
			\hline \hline
			fine-tuned VGG16& 76.68 & 94.06 & 30.93B & 76.68 & 94.06 & 30.93B\\
			\hline
			random selection & 70.25 & 91.16 & 9.63B & 57.28 & 83.52 & 2.62B \\
			ThiNet~\cite{ThiNet} (Our implementation) & 73.00 & 92.27 & 9.63B & 63.12 & 87.54 & 2.62B \\
			\textbf{AutoPruner} (Ours) & \textbf{73.45$\pm$0.26} & \textbf{92.56$\pm$ 0.23} & 9.63B & \textbf{65.06$\pm$0.32} & \textbf{87.93$\pm$0.34} & 2.62B \\
			\hline
		\end{tabular}
	\end{threeparttable}
\end{table*}

We first compare the performance of AutoPruner with others on CUB200-2011~\cite{CUB200}. Many existing model compression algorithms have reported their results on a small dataset like MNIST~\cite{MNIST} or CIFAR-10~\cite{CIFAR}. However, these datasets are relatively simple, and different algorithms often generate very similar results with negligible difference. We argue that comparing on a tough but small dataset is necessary, since it is a more practical application scenario. By contrast, fine-grained recognition is a very challenging task due to the low inter-class but high intra-class variation. 

CUB200-2011 is a popular fine-grained dataset, which aims to recognize 200 bird species. This dataset contains 11,788 bird images. We follow the official train/test split to organize the dataset: 5994 images are used for model training, accuracy will be reported on the rest 5794 images.

\textbf{Implementation details.} We first fine-tune a pretrained VGG16 model on CUB200-2011. For simplicity, only image-level labels are used without other supervised information such as bounding boxes. The images are resized with shorter side=256, then a $224\times 224$ crop is randomly sampled from the resized image with horizontal flip and mean-std normalization. Then the preprocessed images are fed into VGG16 model. We fine-tune VGG16 with 30 epochs using SGD. Weight decay is set to 0.0005, momentum is 0.9 and batch size is set to 64. The initial learning rate starts from 0.001, and is divided by 10 in every 10 epochs. The fine-tuned model achieves $76.683\%$ top-1 accuracy. 

Based on this fine-tuned model, we then train AutoPruner using the same fine-tuning parameters. We prune VGG16 from conv1$\_$1 to conv5$\_$3 layer by layer, \ie, the output of former stage is the input of current stage. At each stage (\eg, we want to prune conv1$\_$1 layer), the AutoPruner module is appended on the output of current layer, and fine-tuned in 2 epochs using the same parameters. The hyper-parameter ($\alpha$) setting is kept the same as what we have stated in Sec.~\ref{sec_implementation} ($\alpha_{start}=0.1, \alpha_{stop}=2$). After fine-tuning, all the filters and channels corresponding to the zeros index values will be pruned away. The new added AutoPruner layer will also be removed. Hence, the only difference after our processing is the reduction of filter number. When pruning is finished on all layers, we will fine-tune the pruned model by another 30 epochs with the same parameters. This pruning pipeline is also applied in other baseline methods for a fair comparison.

\textbf{Comparison among different algorithms.} We compare the proposed AutoPruner method with two approaches:
\begin{itemize}
	\item \textbf{Random selection}. This is a simple but very powerful baseline method. At each pruning stage, several filters are randomly discarded to reduce the complexity of CNN models. As indicated by~\cite{ThiNet}, random selection may be even better than some heuristic methods when compression rate is large. 
	\item \textbf{ThiNet}~\cite{ThiNet}. ThiNet is an efficient three-stage pruning method, which formally establishes filter pruning as an optimization problem and uses statistics of next layer to guide current layer. We re-implement this method with our pruning pipeline for a fair comparison.
\end{itemize}

In order to generate the same network structure among different methods, we first prune the pretrained VGG16 model using AutoPruner. Since the actual compression rate is determined by network itself, AutoPruner may produce slightly larger or smaller network than we expected. Then, according to its output, the same number of filters are pruned away using the above two baseline methods.

Table~\ref{vgg16_CUB} shows the compression results on VGG16 using different filter-level pruning methods. As we can see, AutoPruner is superior over the state-of-the-art filter pruning method ThiNet. Our AutoPruner yields 0.655\% (one result of three repeated experiments) higher top-1 accuracy than ThiNet when $r=0.5$. This advantage will be further expanded when a smaller $r$ is adopted ($r=0.2$), \ie, more filters will be discarded. Since their pruning pipeline and fine-tuning parameters are the same, these two results should reflect that our end-to-end trainable framework is better than previous three-stage pruning method ThiNet. And both of these two models are better than random selection.

\subsection{Ablation Study}
We then conduct ablation studies about the proposed AutoPruner method. This section is composed by two parts: AutoPruner modules and hyper-parameter.

\subsubsection{AutoPruner Modules}
The first ablation study is about AutoPruner itself. We want to explore the influence of different pooling binary codes generation approaches. These two baseline methods are briefly summarized as follows:
\begin{itemize}
	\item \textbf{GAP}. We replace the original max pooling of AutoPruner (see section~\ref{sec_pooling} for more details) with GAP (Global Averaged Pooling) layer and keep other modules unchanged. 
	\item \textbf{Scaling factors}. This baseline is similar to SSS~\cite{sparse_selection}. In AutoPruner, the binary codes are generated from activation tensors. But in SSS, it is produced by a set of end-to-end trainable weights (\ie, scaling factors $\vec{\lambda} \in \mathbb{R}^{C}$). We replace the whole AutoPruner layer with scaling factors, and generate binary codes by $y=\text{sigmoid}(\alpha \vec{\lambda})$.
\end{itemize}

\begin{table}
	\caption{Pruning accuracy (\%) on CUB200-2011 dataset using different algorithms and compression rates.}
	\label{ablation1}
	\footnotesize
	\setlength{\tabcolsep}{4pt}
	\centering
	\begin{threeparttable}
		\begin{tabular}{|l|c|c|c|c|}
			\hline
			\multirow{2}{*}{Method} & \multicolumn{2}{c|}{compression rate $r=0.5$} & \multicolumn{2}{c|}{compression rate $r=0.2$} \\
			\cline{2-5} & top-1 (\%) & \#FLOPs & top-1 (\%) & \#FLOPs \\
			\hline \hline
			GAP & 72.97 & 9.62B & 62.70 &2.69B \\
			scaling factors & 68.66 & 8.20B & 68.14 & \textbf{\underline{6.91B}} \\
			\hline
			\textbf{AutoPruner} & \textbf{73.45} & 9.63B & \textbf{65.06} & 2.62B \\
			\hline
		\end{tabular}
	\end{threeparttable}
\end{table}

Table~\ref{ablation1} shows the results of these two baselines. Experimental settings are the same as section~\ref{exp_cub}. Our max pooling is used for reducing GPU memory consumption. Since GAP can convert the original $1\times C \times H' \times W'$ tensor into a $C$-d vector, it seems to be a better choice. However, a large spatial pooling operation is harmful as shown above. It may discard too much information, hence the coding layers can not generate accurate binary code.

The major difference between AutoPruner and scaling factors is whether pruning information will participate in the training of previous layers or not. AutoPruner learns the binary codes from the output of previous layers, while scaling factors only use a vector to indicate the status of a filter (pruning or not). Hence, the gradient of scaling factors could not be propagated back to previous convolution layers. As illustrated in Table~\ref{ablation1}, this idea is much worse than AutoPruner. It may even fail when compression rate $r$ is small ($r=0.2$). In this situation, model FLOPs can be only reduced to 6.91B. Hence, we believe the gradient information flowing out from channel selection layer is also helpful for previous convolution layer training. It can force network paying more attention to the preserved filters.

\subsubsection{Hyper-parameter Study}
\begin{table}
	\caption{Pruning accuracy (\%) on CUB200-2011 dataset using different choice of $\alpha_{start}$ ($r=0.5$).}
	\label{ablation2}
	\setlength{\tabcolsep}{4.7pt}
	\centering
	\begin{threeparttable}
		\begin{tabular}{|c|c|c|c|c|c|}
			\hline
			$\alpha_{start}$ & 0.01 & 0.05 & 0.1 & 0.5 & 1 \\
			\hline
			top-1 (\%) & 73.438 & 73.438 & 73.662 & 72.644 & 72.541 \\
			\hline
		\end{tabular}
	\end{threeparttable}
\end{table}

In the proposed AutoPruner method, there are mainly two additional parameters: compression ratio $r$ and scaling factor $\alpha$. The compression ratio $r$ is decided by specific tasks as we stated in section~\ref{subsec:sparsity}. This is a predefined value, and is the goal of filter pruning. In general, we should make a tradeoff between model accuracy and inference speed to find an appropriate value for $r$. 

As for $\alpha$, it is decided by $\alpha_{start}$ and $\alpha_{stop}$. As introduced in section~\ref{sec_implementation}, $\alpha_{stop}$ is related to network itself. We can try several numbers to find an appropriate value for $\alpha_{stop}$ that can produce binary codes in Eq~\ref{scaled_sigmoid}. Once this value is determined, it will not be changed again. In our experiments, $\alpha_{stop}=2$ is a fixed choice. Hence, $\alpha_{start}$ is the only hyper-parameter that needs to be tested.

Table~\ref{ablation2} shows the influence of different $\alpha_{start}$ values. Experimental settings are kept the same as section~\ref{exp_cub}. Starting from a relative small value ($\alpha_{start} \le 0.1$), the filters will be gradually erased. Our method will check the value of generated codes and increase $\alpha$ quickly if the code is not converged after several iterations. However, if $\alpha_{start}$ is too large, it may become binary at the first few iterations. In this case, AutoPruner will degenerate to random selection. To sum up, our AutoPruner is really robust for the choice of $\alpha_{start}$ as long as it is not too large. In general, $10\times$ smaller than $\alpha_{end}$ will be a good choice.

\subsection{ImageNet ILSVRC-12} \label{exp_ImageNet}
\begin{table*}
	\caption{Comparison results among several state-of-the-art filter level pruning methods on ImageNet. All the accuracies are tested on validation set using the single view central patch crop. All the FLOPs numbers are calculated by Eq.~\ref{eq:FLOPs} for a fair comparison.}
	\label{model_on_ImageNet}
	\setlength{\tabcolsep}{12pt} 
	\centering
	\small
	\begin{threeparttable}
		\begin{tabular}{|l|c|c|c|c|}
			\hline
			Method & Top-1 Acc.&Top-5 Acc.&\#FLOPs &speed up \tnote{1}\\
			\hline \hline
			Original VGG16 model\tnote{2} & 71.59\% & 90.38\% & 30.94B & 1.00$\times$ \\
			\hline
			\textbf{AutoPruner} & \textbf{69.20\%} & \textbf{88.89\%} & 8.17B & 3.79$\times$ \\
			SSS~\cite{sparse_selection} & 68.53\% & 88.20\% & 7.67B & 4.03$\times$\\
			RNP (3$\times$)~\cite{RNP} & - & 87.58\% & - & 3.00$\times$ \\
			RNP (4$\times$)~\cite{RNP} & - & 86.67\% & - & 4.00$\times$ \\
			Channel Pruning (5$\times$)~\cite{Sunjian_ICCV17}\tnote{3} & 67.80\% & 88.10\% & 7.03B & 4.40$\times$ \\
			Taylor expansion-1~\cite{Molchanov17ICLR} & - & 84.50\% & 8.02B &3.86$\times$ \\
			Taylor expansion-2~\cite{Molchanov17ICLR} & - & 87.00\% & 11.54B &2.68$\times$ \\
			Filter Pruning (impl. by~\cite{Sunjian_ICCV17})~\cite{Li17ICLR} & - & 75.30\% & 7.03B & 4.40$\times$ \\
			
			\hline \hline
			Original ResNet-50 model\tnote{2} & 76.15\% & 92.87\% & 7.72B & 1.00$\times$ \\
			\hline
			\textbf{AutoPruner ($r=0.3$)} & \textbf{73.05\%} & \textbf{91.25\%} & 2.64B & 2.92$\times$ \\
			ThiNet-30~\cite{ThiNet} & 68.42\% & 88.30\% & 2.20B & 3.51$\times$ \\
			\hline
			\textbf{AutoPruner ($r=0.5$)} & \textbf{74.76\%} & \textbf{92.15\%} & 3.76B & 2.05$\times$ \\
			AutoPruner with block pruning ($r=0.5$) & 73.84\% & 91.75\% & 4.30B & 1.80$\times$ \\
			Channel Pruning (2$\times$)~\cite{Sunjian_ICCV17}\tnote{4} & 72.30\% & 90.80\% & 5.22B & 1.48$\times$ \\
			SSS (ResNet-26)~\cite{sparse_selection} & 71.82\% & 90.79\% & 4.00B & 1.93$\times$ \\
			ThiNet-50~\cite{ThiNet} & 71.01\% & 90.02\% & 3.41B & 2.27$\times$ \\
			\hline
		\end{tabular}
		\begin{tablenotes}
			\scriptsize
			\item[1] The speed up ratio is a theoretical value computed by FLOPs. It is fair to be compared in a same model structure.
			\item[2] \url{https://pytorch.org/docs/master/torchvision/models.html}
			\item[3] \url{https://github.com/yihui-he/channel-pruning/releases/tag/channel_pruning_5x}
			\item[4] \url{https://github.com/yihui-he/channel-pruning/releases/tag/ResNet-50-2X}
		\end{tablenotes}
	\end{threeparttable}
\end{table*}

We then compare AutoPruner method with other state-of-the-art approaches on the large scale vision recognition task ImageNet ILSVRC-12~\cite{ImageNet}. 

\textbf{Implementation details.} The fine-tuning settings are similar to those in Sec.~\ref{exp_cub}. We randomly crop a $224\times 224$ input on the resized images, and use the same preprocessing pipelines. Then the model is fine-tuned using SGD with 0.0005 weight decay, 0.9 momentum and 256 batch size. For VGG16, we iteratively prune it layer-by-layer with 3 epochs using $r=0.4$. At each iteration, the $\alpha$ of AutoPruner layer starts from 0.1, and stops at 2. During the first 2 epochs, learning rate is set to 0.001, and is divided by 10 at the third epoch. The pruning procedure stops at conv4\_3 layer. Finally, the whole model is fine-tuned for 30 epochs. 

As for ResNet-50, we follow the setting of ThiNet~\cite{ThiNet} to prune the first two intermediate layers of each residual block. We divide the whole residual blocks into 4 groups, and train multiple AutoPruner layers simultaneously. The initial value of $\alpha$ is set to 1, and stops at 100. At each group, the model is trained 8 epochs with the same parameters as VGG16. We prune ResNet-50 with two compression rate $r=0.5$ and $r=0.3$, and leave the last block uncompressed to obtain a higher accuracy. The compressed model with $r=0.3$ is fine-tuned 30 epochs in the final stage.

Table~\ref{model_on_ImageNet} shows the compression results on ImageNet. For a fair comparison, AutoPruner is only compared with filter level pruning methods. The accuracy is reported using a single view central patch crop: the shorter side is resized to 256, followed by a $224\times 224$ center crop as well as mean-std normalization. We re-calculate the FLOPs by Eq.~\ref{eq:FLOPs}. Hence, the FLOPs values reported here may be different with the original ones (\eg, ResNet~\cite{ResNet}, SSS~\cite{sparse_selection}).

\textbf{VGG16.} We first compare the proposed AutoPruner with other state-of-the-arts on VGG16 model. Among these methods, SSS~\cite{sparse_selection} adopts a very similar technique as ours. In SSS, a scaling factor vector $\vec{\lambda}$ is learned during model training. And all the filters will be removed if their corresponding scaling factors are 0. As we have demonstrated in the ablation study, the gradient information flowing out from channel selection layer is also helpful for previous layers. Hence, AutoPruner can achieve a better result than SSS. 

RNP~\cite{RNP} is another novel method that explore filter pruning beyond the three-stage pipeline. As we can see, the proposed AutoPruner outperforms this method by a large margin. We can achieve a better accuracy even with larger compression ratio (AutoPruner vs. RNP (3$\times$)).

We then compare AutoPruner with channel pruning~\cite{Sunjian_ICCV17}. In this method, the pre-trained model is first pruned by a LASSO regression based method and further processed by 3C approach to get a smaller model. 3C is composed by spatial decomposition~\cite{3C1} and channel decomposition~\cite{3C2}. For a fair comparison, we only report its pruning result. Again, the proposed AutoPruner outperforms this novel three-stage pruning method.

\textbf{ResNet-50.} Similar conclusion can also be acquired on ResNet-50 model. Pruning ResNet is a challenging task due to its compact structure. We follow the same pruning strategy with ThiNet~\cite{ThiNet} but achieve much better accuracy. Channel Pruning (2$\times$)~\cite{Sunjian_ICCV17} introduced a channel sampler layer in the first convolution layer of each residual block to reduce the input width. However, our AutoPruner obtains significantly higher accuracy with a much simpler strategy. The same conclusion is also applicable for SSS~\cite{sparse_selection}. Note that, SSS prune the whole blocks on ResNet, which may leads to larger accuracy drop. For a fair comparison, we also conduct block pruning using AutoPruner. Our method achieves $2\%$ higher top-1 accuracy with similar FLOPs.

\subsection{Consistency of Index Code}

In this part, we want to discuss an interesting question about the consistency of index code. Since the generated binary code is used for model pruning, it should be unique for different input mini-batches. In other words, the output of an AutoPruner layer should be consistent for different images. We empirically demonstrate that the proposed AutoPruner method has such kind of capability.

Let us focus on each channel in the index code, which can be regarded as a classification task. Since the index code is binary, the AutoPruner layer is trained on how to mark all the images with a positive/negative label. This is a relatively simple task. By gradually increasing/decreasing the bias term of fully-connected layer, we can always project all the examples into a positive/negative label. However, if our adaptive sigmoid layer is removed, the network fails to generate a unique output, \ie, one channel is marked useful for some images but useless for others. In AutoPruner, the combination of our mini-batch pooling and binarization ensures the consistency.

To validate this hypothesis, we remove $\alpha$, and train the first group of ResNet-50 using $r=0.3$ on ImageNet. Without increasing $\alpha$ to achieve binarization, the top-1 accuracy of pruned model on validation set is only 9.154\% without fine-tuning. We find that more than 90\% elements in the generated index code are around 0.2. Removing these filters will damage model accuracy greatly since they are not equal to 0. Hence, our novel binarization scheme plays an essential role in the success of AutoPruner.

\subsection{The Value of Network Pruning}
Finally, we will give a brief discussion about the value of network pruning. It is generally accepted that deep model is over-parameterized. This redundancy is helpful for model training, but will significantly slow down inference speed. Hence, there is a great need to remove these redundant parameters after training. 

However, as indicated in the recent study~\cite{rethinking}, training from scratch may achieve even better results than model pruning. For example, they found that if the ThiNet-50 baseline (listed in Table~\ref{model_on_ImageNet}) is trained from scratch by 180 epochs, its top-1 accuracy can be  73.90\%. Although this result is much higher than most existing pruning methods, it is still lower than our AutoPruner (74.76\%). 

To sum up, pruning indeed provides a useful tool to accelerate model inference speed while preserve its accuracy.

\section{Conclusions}

We propose AutoPruner, an end-to-end trainable filter pruning method for CNN acceleration. AutoPruner can be regarded as an independent layer, and can be appended in any convolution layer to prune filters automatically. We demonstrate that the proposed method can significantly improve model compression performance over existing filter pruning methods. Further study reveals that our method can also outperform training from scratch and is really useful to obtain a smaller but still accurate model. In the future, we will study the performance of AutoPruner in other vision tasks, such as object detection or semantic segmentation.

%\clearpage

{\small
\bibliographystyle{ieee}
\bibliography{egbib}

\begin{thebibliography}{10}\itemsep=-1pt

\bibitem{Aghasi17NIPS}
A.~Aghasi, A.~Abdi, N.~Nguyen, and J.~Romberg.
\newblock {Net-Trim: Convex pruning of deep neural networks with performance
  guarantee}.
\newblock In {\em NIPS}, pages 3180--3189, 2017.

\bibitem{Alvarez17NIPS}
J.~Alvarez and M.~Salzmann.
\newblock {Compression-aware training of deep networks}.
\newblock In {\em NIPS}, pages 856--867, 2017.

\bibitem{Chen15ICML}
W.~Chen, J.~Wilson, S.~Tyree, K.~Weinberger, and Y.~Chen.
\newblock {Compressing neural networks with the hashing trick}.
\newblock In {\em ICML}, pages 2285--2294, 2015.

\bibitem{Denil13NIPS}
M.~Denil, B.~Shakibi, L.~Dinh, and N.~de~Freitas.
\newblock {Predicting parameters in deep learning}.
\newblock In {\em NIPS}, pages 2148--2156, 2013.

\bibitem{Denton14NIPS}
E.~L. Denton, W.~Zaremba, J.~Bruna, Y.~LeCun, and R.~Fergus.
\newblock {Exploiting linear structure within convolutional networks for
  efficient evaluation}.
\newblock In {\em NIPS}, pages 1269--1277, 2014.

\bibitem{Han16ICLR}
S.~Han, H.~Mao, and W.~J. Dally.
\newblock {Deep compression: Compressing deep neural networks with pruning,
  trained quantization and huffman coding}.
\newblock In {\em ICLR}, 2016.

\bibitem{Han15NIPS}
S.~Han, J.~Pool, J.~Tran, and W.~Dally.
\newblock {Learning both weights and connections for efficient neural network}.
\newblock In {\em NIPS}, pages 1135--1143, 2015.

\bibitem{MSRA_Init}
K.~He, X.~Zhang, S.~Ren, and J.~Sun.
\newblock {Delving deep into rectifiers: Surpassing human-level performance on
  imagenet classification}.
\newblock In {\em ICCV}, pages 1026--1034, 2015.

\bibitem{ResNet}
K.~He, X.~Zhang, S.~Ren, and J.~Sun.
\newblock {Deep residual learning for image recognition}.
\newblock In {\em CVPR}, pages 770--778, 2016.

\bibitem{AMC}
Y.~He, J.~Lin, Z.~Liu, H.~Wang, L.~Li, and S.~Han.
\newblock {{AMC: AutoML for model compression and acceleration on mobile
  devices}}.
\newblock In {\em ECCV}, volume 11211 of {\em LNCS}, pages 815--832. 2018.

\bibitem{Sunjian_ICCV17}
Y.~He, X.~Zhang, and J.~Sun.
\newblock {Channel pruning for accelerating very deep neural networks}.
\newblock In {\em ICCV}, pages 1389--1397, 2017.

\bibitem{sparse_selection}
Z.~Huang and N.~Wang.
\newblock {{Data-Driven sparse structure selection for deep neural networks}}.
\newblock In {\em ECCV}, volume 11220 of {\em LNCS}, pages 317--334. 2018.

\bibitem{3C1}
M.~Jaderberg, A.~Vedaldi, and A.~Zisserman.
\newblock {Speeding up convolutional neural networks with low rank expansions}.
\newblock In {\em arXiv preprint arXiv:1405.3866}, 2014.

\bibitem{CIFAR}
A.~Krizhevsky.
\newblock Learning multiple layers of features from tiny images.
\newblock Master's thesis, University of Toronto, 2009.

\bibitem{Lebedev16CVPR}
V.~Lebedev and V.~Lempitsky.
\newblock {Fast convnets using group-wise brain damage}.
\newblock In {\em CVPR}, pages 2554--2564, 2016.

\bibitem{MNIST}
Y.~LeCun, L.~Bottou, Y.~Bengio, and P.~Haffner.
\newblock {Gradient-based learning applied to document recognition}.
\newblock {\em Proceedings of the IEEE}, 86(11):2278--2324, 1998.

\bibitem{Li17ICLR}
H.~Li, A.~Kadav, I.~Durdanovic, H.~Samet, and H.~P. Graf.
\newblock {Pruning filters for efficient ConvNets}.
\newblock In {\em ICLR}, 2017.

\bibitem{RNP}
J.~Lin, Y.~Rao, J.~Lu, and J.~Zhou.
\newblock {Runtime neural pruning}.
\newblock In {\em NIPS}, pages 2178--2188, 2017.

\bibitem{JianguoLi_ICCV17}
Z.~Liu, J.~Li, Z.~Shen, G.~Huang, S.~Yan, and C.~Zhang.
\newblock {Learning efficient convolutional networks through network slimming}.
\newblock In {\em ICCV}, pages 2755--2763, 2017.

\bibitem{rethinking}
Z.~Liu, M.~Sun, T.~Zhou, G.~Huang, and T.~Darrell.
\newblock {Rethinking the Value of Network Pruning}.
\newblock In {\em arXiv preprint arXiv:1810.05270}, 2018.

\bibitem{Louizos17NIPS}
C.~Louizos, K.~Ullrich, and M.~Welling.
\newblock {Bayesian compression for deep learning}.
\newblock In {\em NIPS}, pages 3290--3300, 2017.

\bibitem{ThiNet}
J.~Luo, J.~Wu, and W.~Lin.
\newblock {ThiNet: A filter level pruning method for deep neural network
  compression}.
\newblock In {\em ICCV}, pages 5058--5066, 2017.

\bibitem{Molchanov17ICLR}
P.~Molchanov, S.~Tyree, T.~Karras, T.~Aila, and J.~Kautz.
\newblock {Pruning convolutional neural networks for resource efficient
  inference}.
\newblock In {\em ICLR}, 2017.

\bibitem{ImageNet}
O.~Russakovsky, J.~Deng, H.~Su, J.~Krause, S.~Satheesh, S.~Ma, Z.~Huang,
  A.~Karpathy, A.~Khosla, M.~Bernstein, A.~C. Berg, and F.-F. Li.
\newblock {ImageNet large scale visual recognition challenge}.
\newblock {\em IJCV}, 115(3):211--252, 2015.

\bibitem{VGG16}
K.~Simonyan and A.~Zisserman.
\newblock {Very deep convolutional networks for large-scale image recognition}.
\newblock In {\em ICLR}, 2015.

\bibitem{CUB200}
C.~Wah, S.~Branson, P.~Welinder, P.~Perona, and S.~Belongie.
\newblock {The Caltech-UCSD birds-200-2011 dataset}.
\newblock Technical Report CNS-TR-2011-001, California Institute of Technology,
  2011.

\bibitem{Wen18ICLR}
W.~Wen, Y.~He, S.~Rajbhandari, W.~Wang, F.~Liu, B.~Hu, Y.~Chen, and H.~Li.
\newblock {Learning intrinsic sparse structures within long short-term memory}.
\newblock In {\em ICLR}, 2018.

\bibitem{Wen16NIPS}
W.~Wen, C.~Wu, Y.~Wang, Y.~Chen, and H.~Li.
\newblock {Learning structured sparsity in deep neural networks}.
\newblock In {\em NIPS}, pages 2074--2082, 2016.

\bibitem{Wu16CVPR}
J.~Wu, C.~Leng, Y.~Wang, Q.~Hu, and J.~Cheng.
\newblock {Quantized convolutional neural networks for mobile devices}.
\newblock In {\em CVPR}, pages 4820--4828, 2016.

\bibitem{NISP}
R.~Yu, A.~Li, C.~Chen, J.~Lai, V.~Morariu, X.~Han, M.~Gao, C.~Lin, and
  L.~Davis.
\newblock {NISP: Pruning networks using neuron importance score propagation}.
\newblock In {\em CVPR}, 2018.

\bibitem{3C2}
X.~Zhang, J.~Zou, K.~He, and J.~Sun.
\newblock Accelerating very deep convolutional networks for classification and
  detection.
\newblock {\em TPAMI}, 38(10):1943--1955, 2016.

\end{thebibliography}
}

\end{document}